\documentclass[conference]{IEEEtran}
\IEEEoverridecommandlockouts

\usepackage{cite}
\usepackage{tikz}
\usetikzlibrary{positioning, shapes, arrows.meta}
\usepackage{amsmath,amssymb,amsfonts}
\usepackage{algorithmic}
\usepackage{graphicx}
\usepackage{textcomp}
\usepackage{xcolor}
\usepackage{multirow}
\usepackage{afterpage}
\def\BibTeX{{\rm B\kern-.05em{\sc i\kern-.025em b}\kern-.08em
    T\kern-.1667em\lower.7ex\hbox{E}\kern-.125emX}}
\begin{document}

\title{A Pointcloud Registration Framework for Relocalization in Subterranean Environments}

\author{David Akhihiero$^{1}$ and Jason N. Gross$^{1}$
\thanks{$^{1}$Department of Mechanical, Materials and Aerospace Engineering, West Virginia University}
}

\maketitle

\section*{Abstract}
Relocalization, the process of re-establishing a robot's position within an environment, is crucial for ensuring accurate navigation and task execution when external positioning information, such as GPS, is unavailable or has been lost. Subterranean environments present significant challenges for relocalization due to limited external positioning information, poor lighting that affects camera localization, irregular and often non-distinct surfaces, and dust, which can introduce noise and occlusion in sensor data. In this work, we propose a robust, computationally friendly framework for relocalization through point cloud registration utilizing a prior point cloud map. The framework employs Intrinsic Shape Signatures (ISS) to select feature points in both the target and prior point clouds. The Fast Point Feature Histogram (FPFH) algorithm is utilized to create descriptors for these feature points, and matching these descriptors yields correspondences between the point clouds. A 3D transformation is estimated using the matched points, which initializes a Normal Distribution Transform (NDT) registration. The transformation result from NDT is further refined using the Iterative Closest Point (ICP) registration algorithm. This framework enhances registration accuracy even in challenging conditions, such as dust interference and significant initial transformations between the target and source, making it suitable for autonomous robots operating in underground mines and tunnels. This framework was validated with experiments in simulated and real-world mine datasets, demonstrating its potential for improving relocalization. The contributions of this work are: 1) a robust framework for relocalization in challenging subterranean environments, addressing noise, occlusions, and irregular surfaces with a multi-stage registration process; 2) a computationally efficient approach, integrating ISS keypoint selection, FPFH descriptors, and NDT initialization to support real-time operations; and 3) validation of the framework on simulated and real-world mine datasets, demonstrating practical applicability for autonomous navigation in underground settings.

\section{Introduction}
Autonomous robotic systems are increasingly in demand in complex and unstructured environments, especially those requiring routine inspections and infrastructure monitoring. \cite{sayed2022modular, azpurua2021towards}. In subterranean environments like underground mines and tunnels, robots are used for exploration \cite{mascarich2018multi}, search, and rescue \cite{wang2014development}, and safety inspections \cite{tang2019novel,szrek2022mobile, martinez2023oxpecker}. These environments present unique challenges like limited visibility due to poor lighting, dust, and debris that can obstruct sensor data, irregular and non-distinct surfaces, rough terrains, and risk of collapse \cite{akhihiero2024cooperative, zimroz2019should}. A critical aspect of robot navigation in these environments is the ability to relocalize after losing track of its position. Another equally important challenge is accurate and efficient mapping, which serves as a foundation for navigation and decision-making. Mapping in subterranean settings can be improved by careful trajectory design \cite{samarakoon2022impact, SCC.Ogri.Qureshi.ea2024} to increase feature detection and feature overlap.

In scenarios where the robot has previously mapped part of the environment and needs to map other sections or needs to remap the same section for inspection purposes, it needs to be able to relocalize itself within the old map. In the scenario where the old map is stored as a pointcloud, one way for the robot to relocalize is to start building a new pointcloud map and then register the new pointcloud with the old one. There are multiple potential registration strategies; one approach is to use the Iterative Closest Point (ICP) algorithm \cite{li2020evaluation} or its variants to estimate the transformation between the two point clouds. The problem with ICP is that the result depends strongly on the initial guess for the transformation and its solution can converge to a local minimum \cite{pomerleau2013comparing}. It is also computationally expensive but it will converge to the correct solution if it has a good initial guess. The Normal Distributions Transform (NDT) algorithm \cite{magnusson2007scan} is often faster and more robust to poor initial transformations and noise than the ICP \cite{magnusson2009evaluation} since it represents the point clouds as a set of Gaussian distributions in a grid rather than individual points but its result is still dependent on the initial transformation and in several cases, it will converge to the wrong local minimum. Another registration strategy is to obtain correspondences between the target and the prior point clouds and estimate a 3D transformation to align both point clouds using these correspondences. Correspondences are obtained by matching feature descriptors in both point clouds and these descriptors are typically generated on keypoints extracted from each point cloud. To address the challenge of estimating a correct transformation with bad correspondences, several transformation estimation algorithms have been proposed like the random sample consensus (RANSAC) \cite{martinez2022ransac}, sample consensus initial alignment (SAC-IA) \cite{rusu2009fast}, game theoretical matching (GTM) \cite{albarelli2015fast}. 3D keypoints can be detected using the Local Surface Patches (LSP) \cite{chen20073d}, Intrinsic Shape Signatures (ISS) \cite{zhong2009intrinsic}, Keypoint Quality (KPQ) \cite{mian2010repeatability}, Laplace-Beltrami Scale Space (LBSS)\cite{unnikrishnan2008multi} and even deep-learning based methods \cite{liu2022rethinking}. ISS keypoints are reasonably repeatable, selection is very efficient \cite{tombari2013performance} and there are easy-to-use open-source implementations of this algorithm in the point cloud library \cite{rusu20113d} and Open3D \cite{zhou2018open3d}. There are several options for computing feature descriptors for keypoints like the Point Feature Histogram (PFH) \cite{rusu2008persistent}, Fast Point Feature Histogram (FPFH) \cite{rusu2009fast}, Signature of Histograms of OrienTations (SHOT) \cite{salti2014shot}. The FPFH descriptor which was developed from the PFH algorithm has a lower computational complexity and achieves the best results in most cases \cite{szalai2024fpfh}. The SHOT descriptor is more robust to noise \cite{guo2015performance} but is more computationally expensive and is sometimes unstable \cite{kim2013evaluation}.

In this work,  a relocalization framework for subterranean environments, employing the Fast Point Feature Histogram (FPFH) \cite{rusu2009fast} algorithm to compute descriptors for Intrinsic Shape Signatures (ISS) \cite{zhong2009intrinsic} keypoints in both source and target point clouds is proposed and evaluated. The proposed approach begins by obtaining point correspondences through descriptor matching between the source point cloud map and the target point cloud. An initial 3D geometric transformation based on these correspondences is then estimated. To mitigate the impact of noise and false correspondences, the transformation estimate is refined in two stages. First, the Normal Distributions Transform (NDT) \cite{magnusson2007scan} registration, which is robust to noise, is applied followed by Iterative Closest Point (ICP) \cite{li2020evaluation} registration to fine-tune the final transformation. This two-step refinement enhances the accuracy and reliability of relocalization, aligning with our contributions of providing a robust framework to address noise, occlusions, and irregular surfaces, along with a computationally efficient approach suitable for real-time operations. It is uncommon for pointcloud registration to include two refinement stages. Most registration approaches have just two stages (one coarse registration stage and then a refinement stage) \cite{geng2024improved, pei2024study}. Our work attempts to show the value of two refinement stages in improving reliability in challenging cases. The addition of NDT before ICP could help mitigate errors that could arise from poor initial alignment, especially in challenging underground environments.

\section{Methodology}
\label{method}
\begin{figure*}[h!]
\resizebox{\textwidth}{!}{  
\begin{tikzpicture}[
    node distance=2cm and 1.5cm,
    every node/.style={rectangle, rounded corners, draw=black, fill=gray!20, text centered, minimum height=2.5cm, minimum width=4cm, inner sep=0pt, text width=4cm}, 
    every arrow/.style={-{Latex[length=3mm]}, thick},
    fancy block/.style={rectangle, rounded corners, draw=blue!70, fill=blue!10, thick, text centered, minimum height=3cm, minimum width=5cm, font=\small, text width=4cm} 
]

\node (source_denoise) [fancy block] {Source Point Cloud \\ Denoised and Downsampled};
\node (source_keypoint) [right=of source_denoise, fancy block] {ISS Keypoints \\ Selected in Source};
\node (source_descriptor) [right=of source_keypoint, fancy block] {FPFH Descriptors Computed \\ for Source Keypoints};

\node (target_denoise) [below=of source_denoise, xshift=0cm, fancy block] {Target Point Cloud \\ Denoised and Downsampled};
\node (target_keypoint) [right=of target_denoise, fancy block] {ISS Keypoints \\ Selected in Target};
\node (target_descriptor) [right=of target_keypoint, fancy block] {FPFH Descriptors Computed \\ for Target Keypoints};

\node (match) [below=of source_descriptor, below=of target_descriptor, right=of source_descriptor, xshift=2cm, fancy block] {Descriptors Matched};

\node (merge) [below=of match, xshift=0cm, fancy block] {Transformation Estimated \\ from Matched Keypoints};
\node (ndt) [below=of merge, fancy block] {NDT Applied to Refine Estimate};
\node (icp) [below=of ndt, fancy block] {ICP Applied to Refine Estimate};

\draw[every arrow] (source_denoise) -- (source_keypoint);
\draw[every arrow] (source_keypoint) -- (source_descriptor);
\draw[every arrow] (source_descriptor) -- (match);

\draw[every arrow] (target_denoise) -- (target_keypoint);
\draw[every arrow] (target_keypoint) -- (target_descriptor);
\draw[every arrow] (target_descriptor) -- (match);

\draw[every arrow] (match) -- (merge);
\draw[every arrow] (merge) -- (ndt);
\draw[every arrow] (ndt) -- (icp);

\end{tikzpicture}
}
\caption{Flowchart of the relocalization framework for subterranean environments, illustrating the steps of denoising, downsampling, keypoint selection, descriptor computation, and transformation refinement.}
\label{relocalization_flowchart}
\end{figure*}
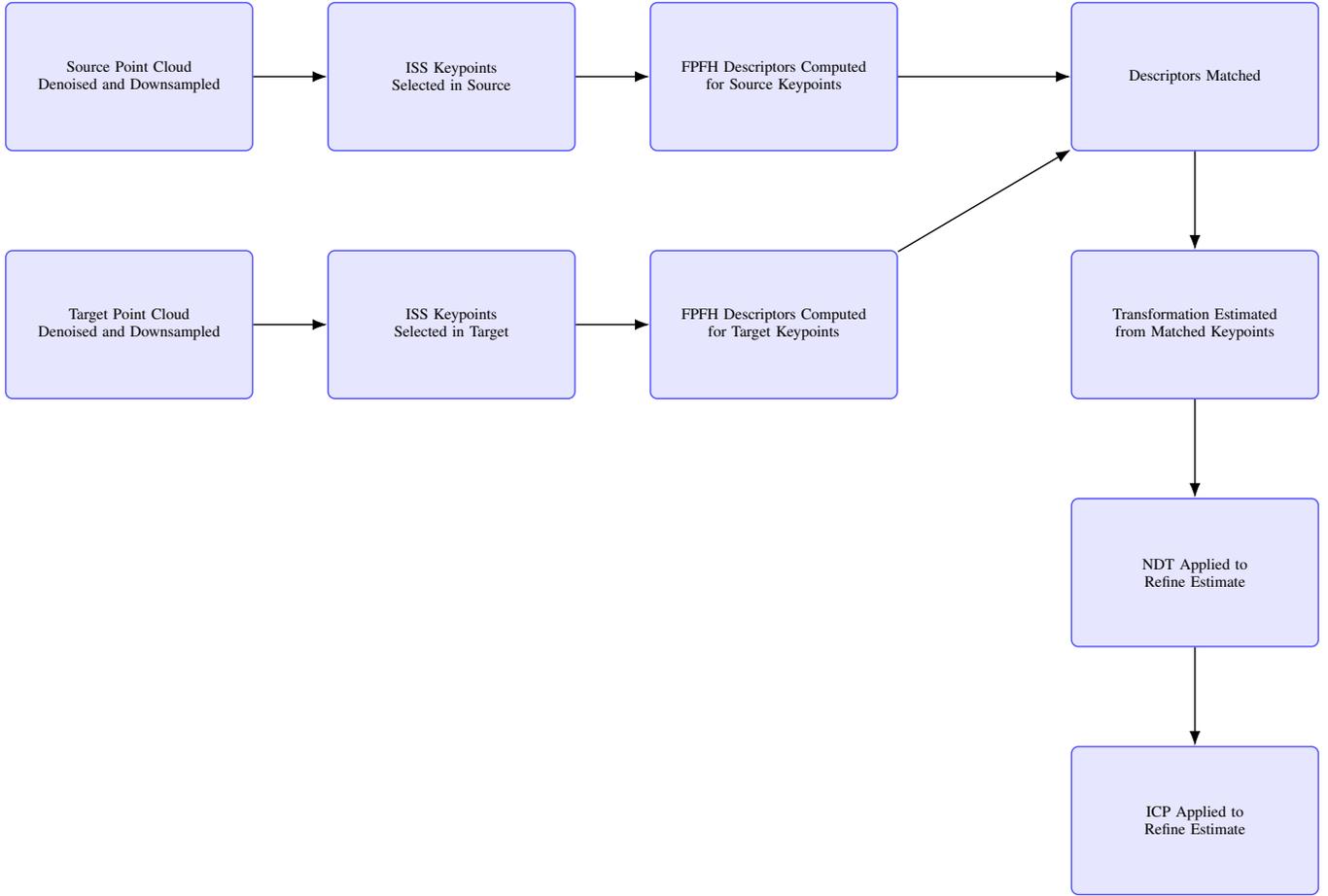

This section outlines the procedure for the proposed relocalization framework, illustrated in Figure \ref{relocalization_flowchart}. The source pointcloud is the previous map cloud of the environment and the target is the new map cloud. The framework comprises three primary stages: FPFH Descriptor Extraction and Transformation Estimation, Transformation Refinement Using NDT, and Final Alignment Refinement with ICP. Each stage is described below.

\subsection{FPFH Descriptor Extraction and Transformation Estimation}
In this section, the processes involved in extracting Fast Point Feature Histograms (FPFH) \cite{rusu2009fast} from keypoints in the source and target point clouds and estimating the initial transformation that aligns these point clouds based on the extracted features are detailed.

The Intrinsic Shape Signatures (ISS)\cite{zhong2009intrinsic} algorithm is leveraged in the work to extract keypoints with large 3D variations in their local neighborhood. This variation, for a keypoint $p$, is computed using the smallest eigenvalue ($\lambda_1$) of the scatter matrix of the points in a spherical region of radius $r_{salient}$ around $p$. To exclude points where the scatter matrix has two similar eigenvalues, another constraint is imposed on the ratios of the eigenvalues as shown in Equation \ref{iss_selection} \cite{zhong2009intrinsic}. An additional constraint is that at most one keypoint is extracted from a cubic volume of size $d_{voxel}$. An extracted keypoint from this volume is the one with the maximum saliency, where saliency is the magnitude of its smallest eigenvalue.

\begin{equation}
    \begin{split}
        \mathbf{S}_i &= \sum_{j \in N(i)} (\mathbf{p}_j - \bar{\mathbf{p}})(\mathbf{p}_j - \bar{\mathbf{p}})^T, \\
        &\lambda_1 \leq \lambda_2 \leq \lambda_3, \quad \lambda_1 > \lambda_{\text{threshold}}, \\
        &\frac{\lambda_2}{\lambda_1} < \gamma_{21}, \quad \frac{\lambda_3}{\lambda_2} < \gamma_{32}.
    \end{split}
    \label{iss_selection}
\end{equation}

Where,
\begin{itemize}
    \item $\mathbf{S}_i$: Scatter matrix for point $i$.
    \item $N(i)$: Set of neighboring points within a spherical radius $r_{\text{salient}}$ of point $i$.
    \item $\mathbf{p}_j$: Position of the $j$-th neighbor.
    \item $\bar{\mathbf{p}}$: Centroid of the neighboring points around point $i$.
    \item $\lambda_1, \lambda_2, \lambda_3$: Eigenvalues of the scatter matrix $\mathbf{S}_i$, ordered such that $\lambda_1 \leq \lambda_2 \leq \lambda_3$.
    \item $\lambda_{\text{threshold}}$: Threshold for selecting salient points based on $\lambda_1$.
    \item $\gamma_{21}$: Threshold for the ratio of the second to the first eigenvalue, $\frac{\lambda_2}{\lambda_1}$, to avoid ambiguous axes.
    \item $\gamma_{32}$: Threshold for the ratio of the third to the second eigenvalue, $\frac{\lambda_3}{\lambda_2}$, to avoid ambiguous axes.
\end{itemize}

To compute the FPFH descriptor for a keypoint $p_i$ with a normal vector $\mathbf{n}_i$, the neighboring points within a given radius $r$ are selected. A Simplified Point Feature Histogram (SPFH) for each point $p_j$ is computed in its Darboux $uvw$ frame and after computing the SPFH for all neighboring points, their histograms are accumulated. The FPFH for $p_i$ is formed as a weighted combination of its SPFH and the SPFH of all its neighbors giving more weight to nearby points \cite{rusu2009fast}.

\begin{equation*}
u = \mathbf{n}_i, \quad v = (\mathbf{p}_j - \mathbf{p}_i) \times u, \quad w = u \times v
\end{equation*}

\begin{equation*}
\alpha = v \cdot \mathbf{n}_j
\end{equation*}

\begin{equation*}
\varphi = \frac{u \cdot (\mathbf{p}_j - \mathbf{p}_i)}{\|\mathbf{p}_j - \mathbf{p}_i\|}
\end{equation*}

\begin{equation*}
\theta = \arctan\left(w \cdot \mathbf{n}_j, u \cdot \mathbf{n}_j\right)
\end{equation*}

Where,
\begin{itemize}
    \item $\mathbf{n}_i$: the normal at point $\mathbf{p}_i$, with $\mathbf{p}_j$ as a neighboring point.
     \item $\mathbf{n}_j$: the normal at point $\mathbf{p}_j$.
    \item $\alpha$: the angle variation between $v$ and the normal $\mathbf{n}_j$ of neighbor $\mathbf{p}_j$.
    \item $\varphi$: the angle between $u$ and the vector from $\mathbf{p}_i$ to $\mathbf{p}_j$.
    \item $\theta$: the angle between the normal $\mathbf{n}_j$ and the vector $w$.
\end{itemize}

\begin{equation}
FPFH(p_i) = SPFH(p_i) + \frac{1}{k} \sum_{i=1}^{k} \frac{1}{d(p_i, p_k)} \cdot SPFH(p_k)
\label{fpfh_eqn}
\end{equation}

where: \\
$FPFH(p_i)$ is the Fast Point Feature Histogram at point $p_i$, $k$ is the number of neighbors in the k-neighborhood, and $d(p_i, p_k)$ is the Euclidean distance between points $\mathbf{p}_i$ and $\mathbf{p}_k$. Keypoints in the source and target pointclouds are matched based on the similarity of their descriptors using a Euclidean distance metric and the Random Sample Consensus (RANSAC) \cite{martinez2022ransac} is used to estimate the transformation between matched keypoints.

\subsection{Transformation Refinement Using NDT}
In this section, how the Normal Distributions Transform (NDT) \cite{jun2015point} is employed to refine the transformation estimated from the FPFH correspondences is described. The source pointcloud is first discretized into a 3D voxel grid and within each voxel, a normal distribution is estimated based on the points in that voxel. The mean and covariance represent the local point distribution as shown in Equation \ref{ndt_eqn}\cite{jun2015point}.

\begin{equation}
N(\mathbf{x}; \boldsymbol{\mu}, \Sigma) = \frac{1}{(2\pi)^{\frac{k}{2}} |\Sigma|^{\frac{1}{2}}} \exp\left(-\frac{1}{2}(\mathbf{x} - \boldsymbol{\mu})^T \Sigma^{-1} (\mathbf{x} - \boldsymbol{\mu})\right)
\label{ndt_eqn}
\end{equation}
\text{where:}
\begin{itemize}
    \item $\mathbf{x}$ is a point in the voxel.
    \item $\boldsymbol{\mu}$ is the mean of the distribution (centroid of points in the voxel).
    \item $\Sigma$ is the covariance matrix of the points in the voxel.
    \item $k$ is the dimensionality of the points (in 3D, $k=3$).
\end{itemize}

The target pointcloud is transformed using the initial transformation estimate and for each point in the transformed pointcloud, the likelihood of that point given the normal distributions defined in the source voxel grid is evaluated. Gradient descent optimization is used, iteratively, to maximize the likelihood of the target pointcloud fitting the source's normal distributions.

\subsection{Final Alignment Refinement with ICP}
After the NDT alignment, the Iterative Closest Point (ICP) algorithm \cite{li2020evaluation} is used to refine the transformation between the source and target point clouds by minimizing the error between corresponding points. For each point, $p_i$, in the target pointcloud, the closest corresponding point, $p_j$, in the source cloud is formed. This correspondence is updated at each iteration. The ICP algorithm minimizes the cost function in Equation \ref{icp}\cite{li2020evaluation}.

\begin{equation}
\label{icp}
\mathcal{L}_{ICP} = \sum_{i=1}^{N} \left\| T(p_i) - p_j \right\|^2
\end{equation}

where $T(p_i)$ is the transformed position of point $p_i$, $p_j$ is the closest corresponding point in the source cloud, and $N$ is the number of points in the target cloud.

\section{Results}
\label{experimental_results}

\begin{figure}[h!]
    \centering
    \includegraphics[width=0.49\linewidth]{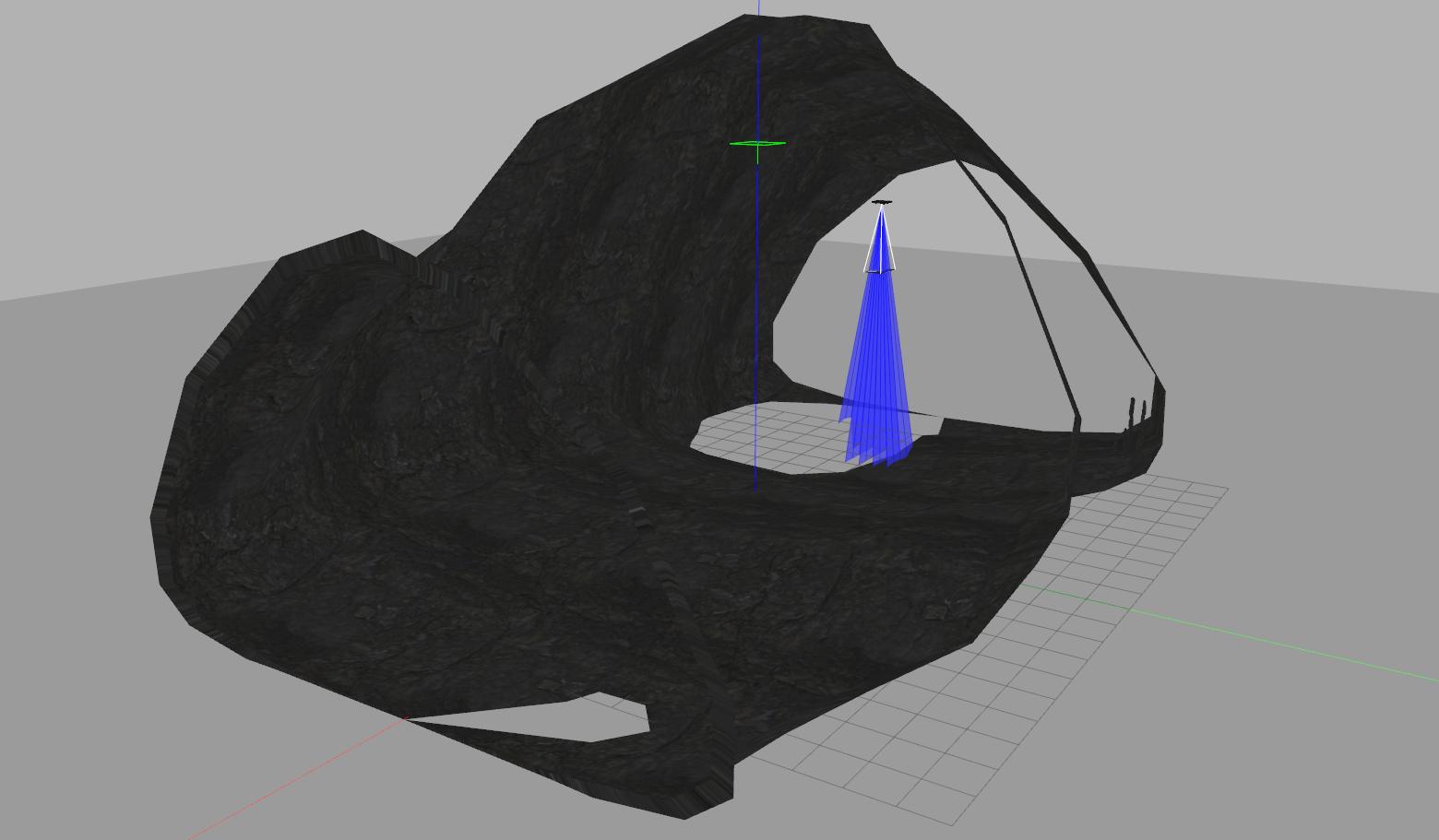} 
    \includegraphics[width=0.49\linewidth, trim=0 50 0 72, clip]{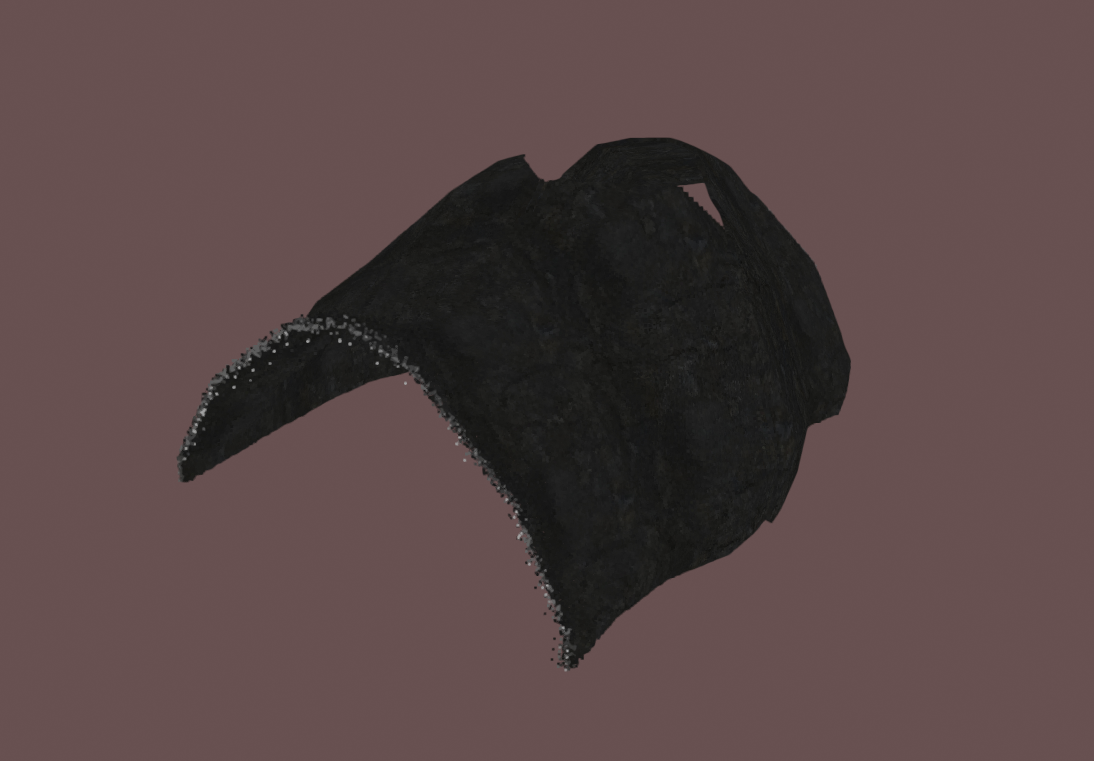} 
    \textbf{Gazebo simulator} \hspace{4em} \textbf{Ground-truth pointcloud} 
    \caption{Model mine environment used in the ROS/Gazebo \cite{takaya2016simulation} simulator.}
    \label{gazebo_mine}
\end{figure}

\begin{figure*}[p]
    \centering
    \begin{tabular}{cc}
        \includegraphics[width=0.49\textwidth, clip, trim=300 100 300 150]{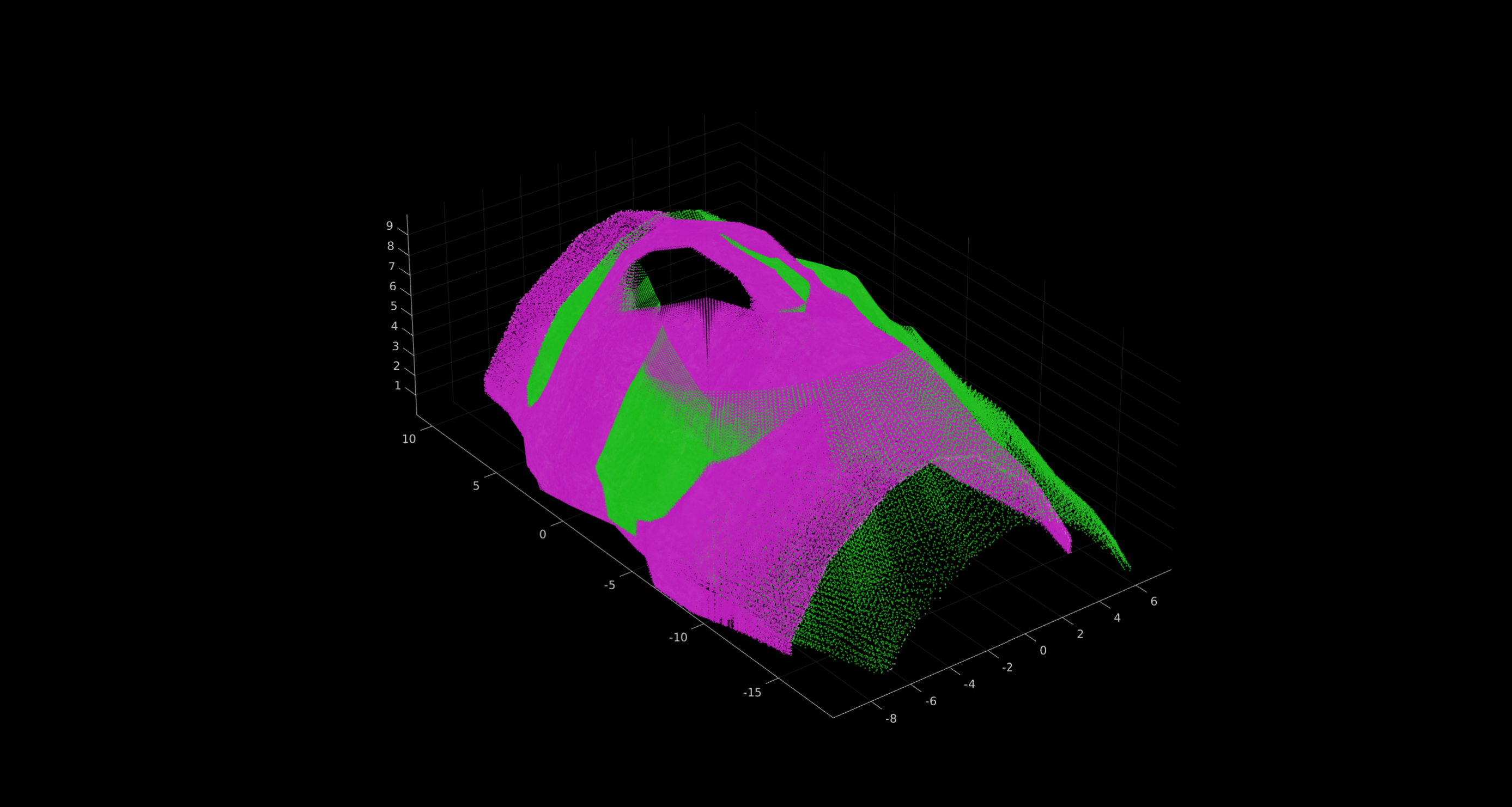} & 
        \includegraphics[width=0.49\textwidth, clip, trim=300 100 300 150]{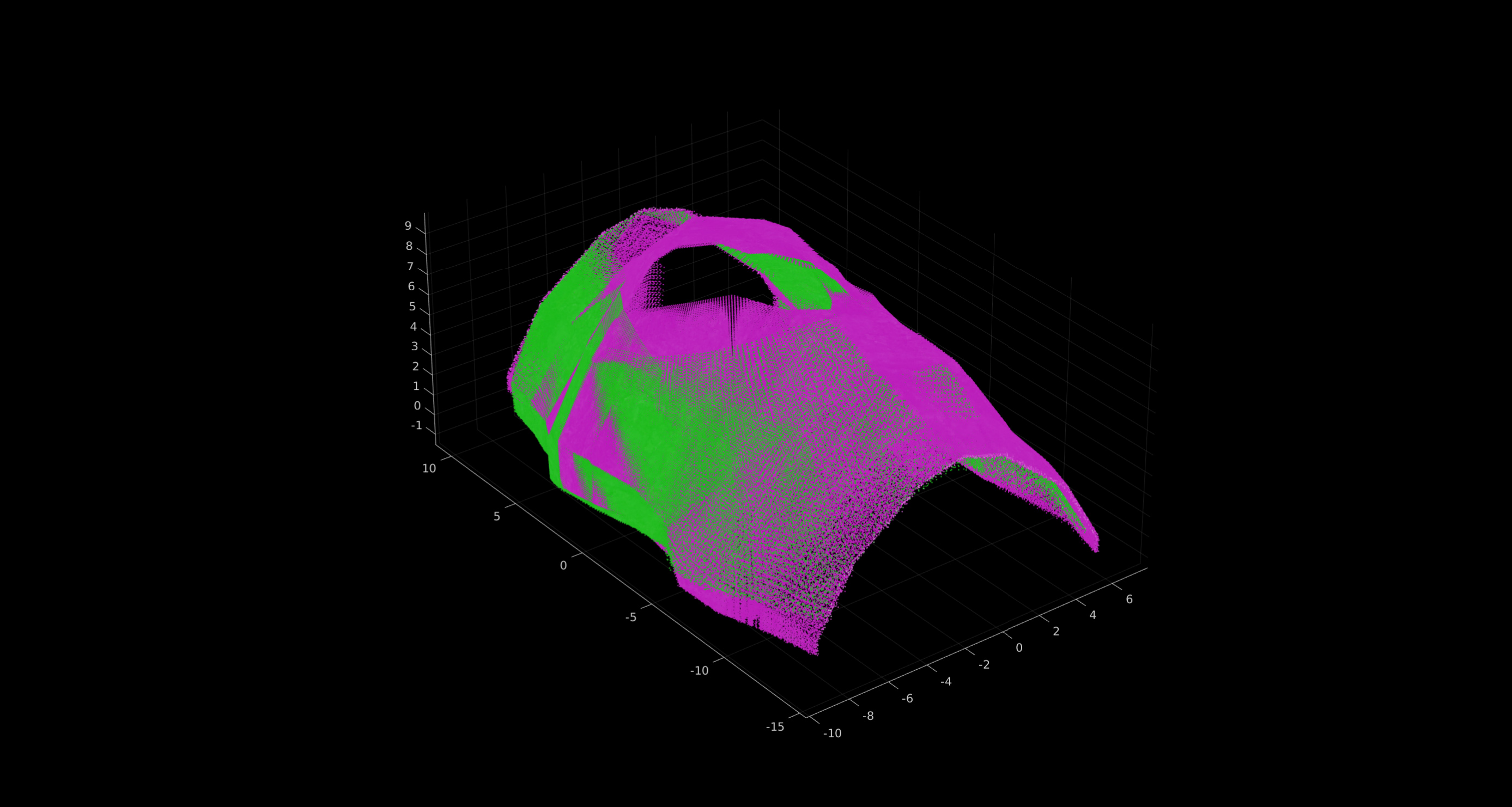} \\
        \textbf{Pointcloud 1 (Before Registration)} & \textbf{Pointcloud 1 (After Registration)} \\
        
        \includegraphics[width=0.49\textwidth, clip, trim=300 100 300 150]{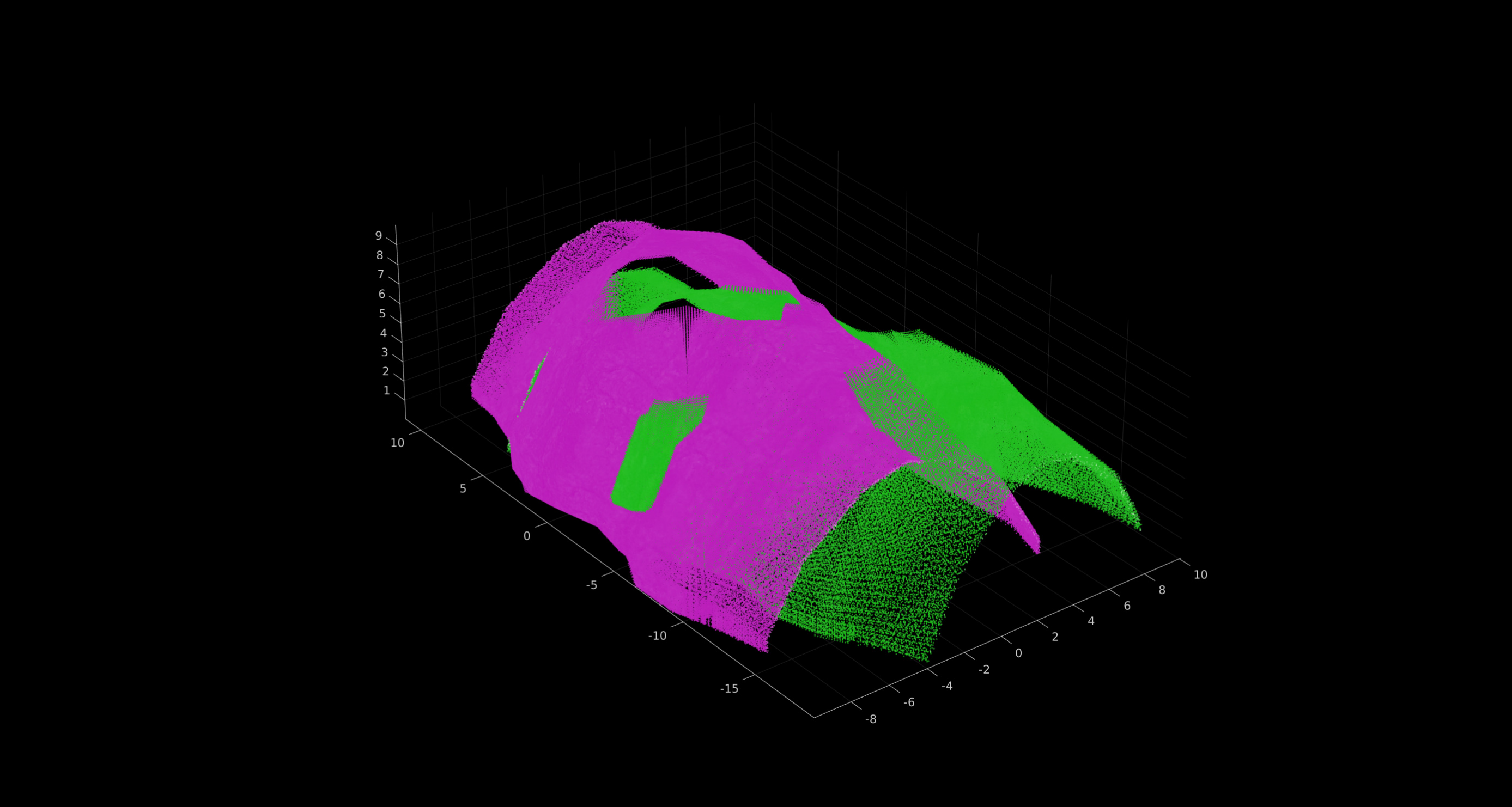} &
        \includegraphics[width=0.49\textwidth, clip, trim=300 100 300 150]{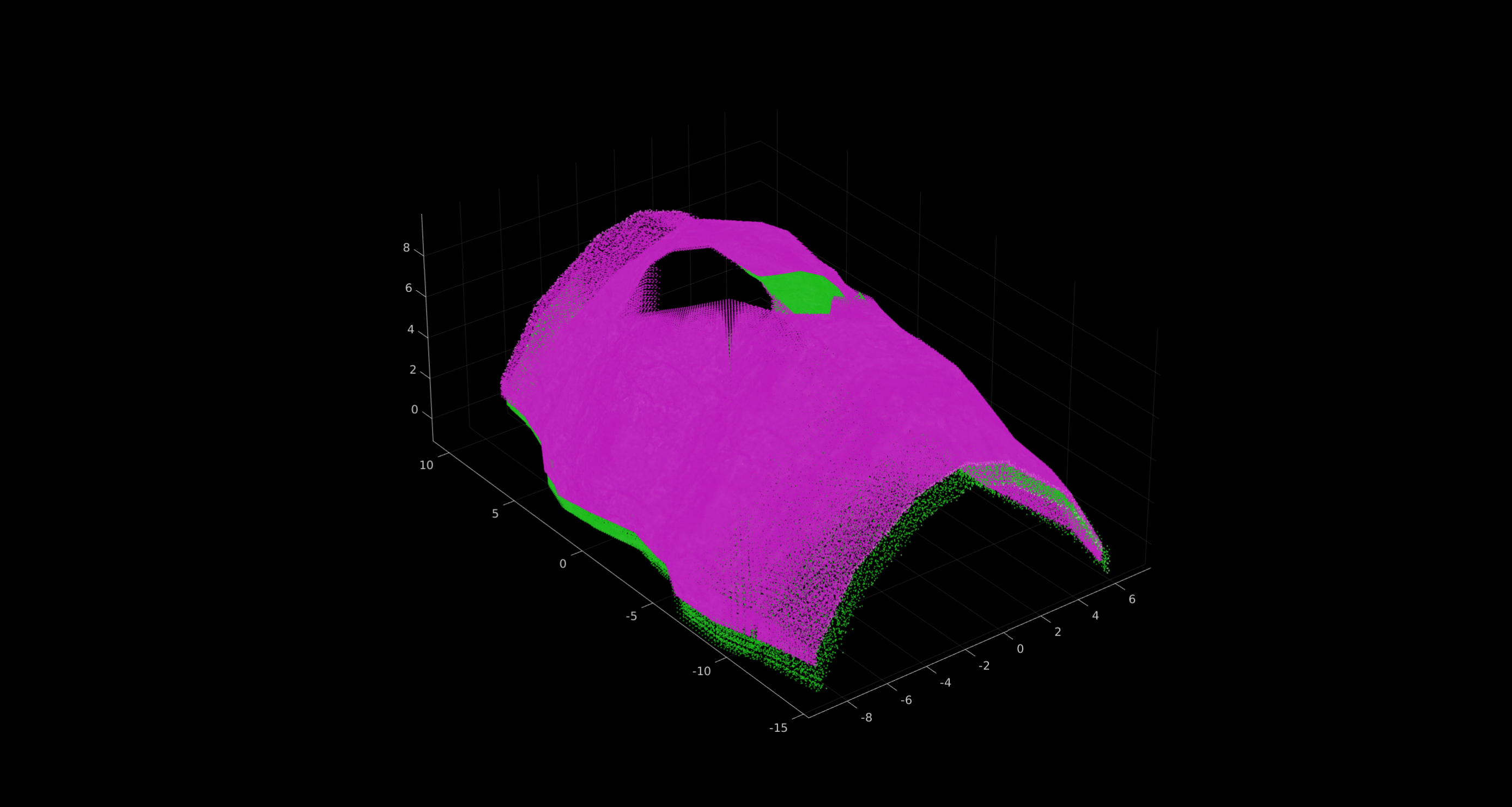} \\
        \textbf{Pointcloud 2 (Before Registration)} & \textbf{Pointcloud 2 (After Registration)} \\
        
        \includegraphics[width=0.49\textwidth, clip, trim=300 100 300 150]{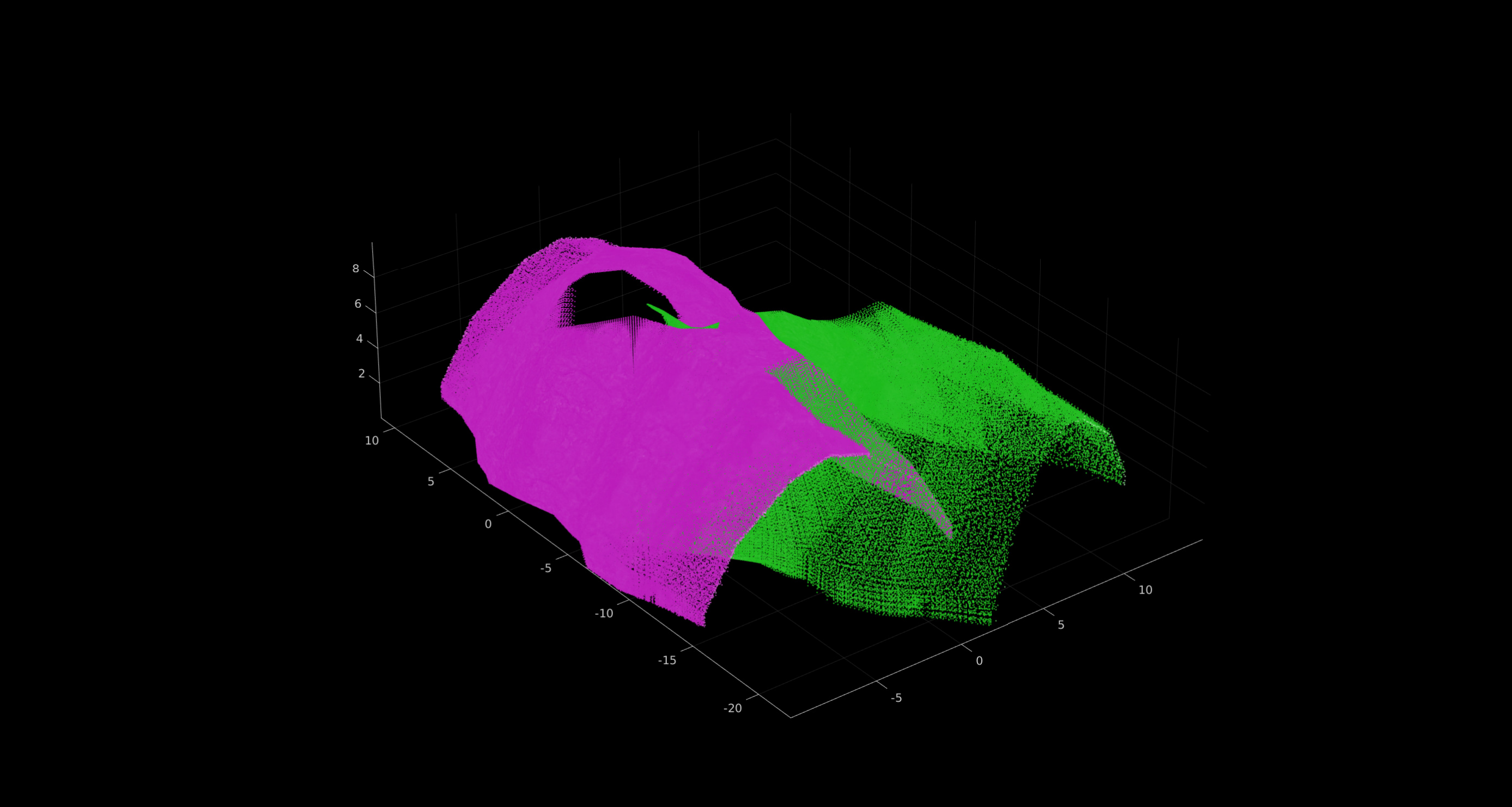} &
        \includegraphics[width=0.49\textwidth, clip, trim=300 100 300 150]{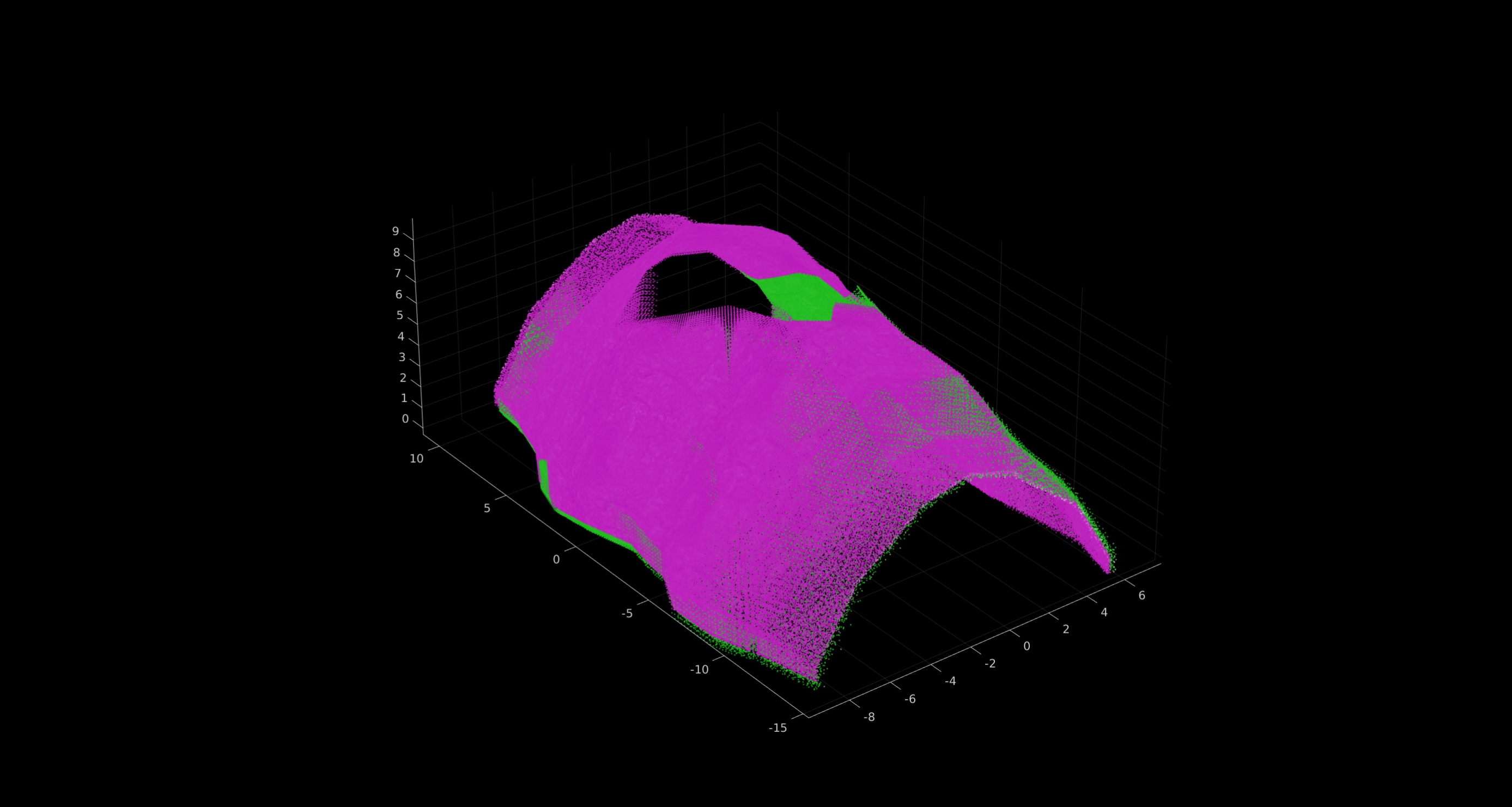} \\
        \textbf{Pointcloud 3 (Before Registration)} & \textbf{Pointcloud 3 (After Registration)} \\
        
        \includegraphics[width=0.49\textwidth, clip, trim=300 100 300 150]{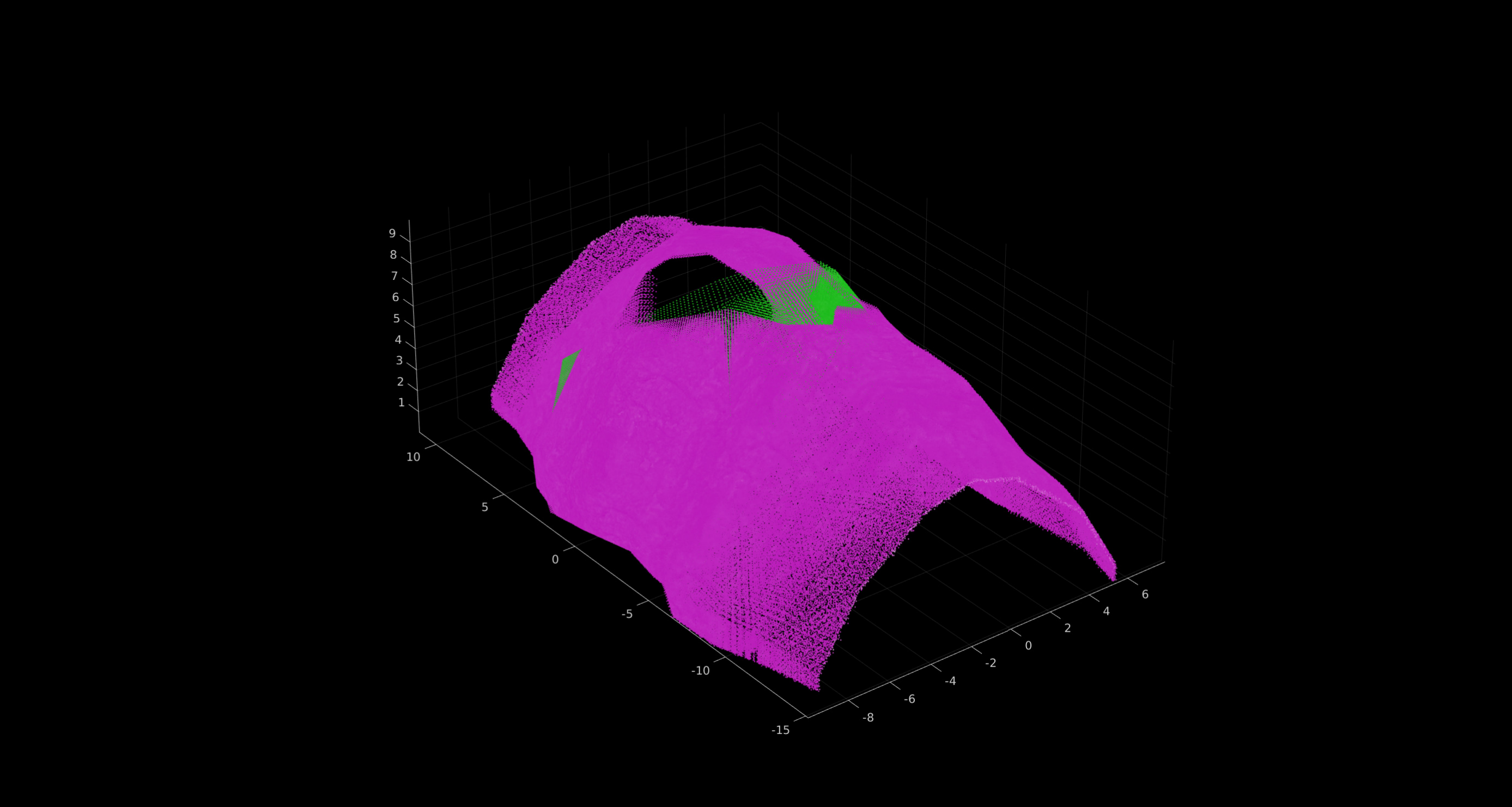} &
        \includegraphics[width=0.49\textwidth, clip, trim=300 100 300 150]{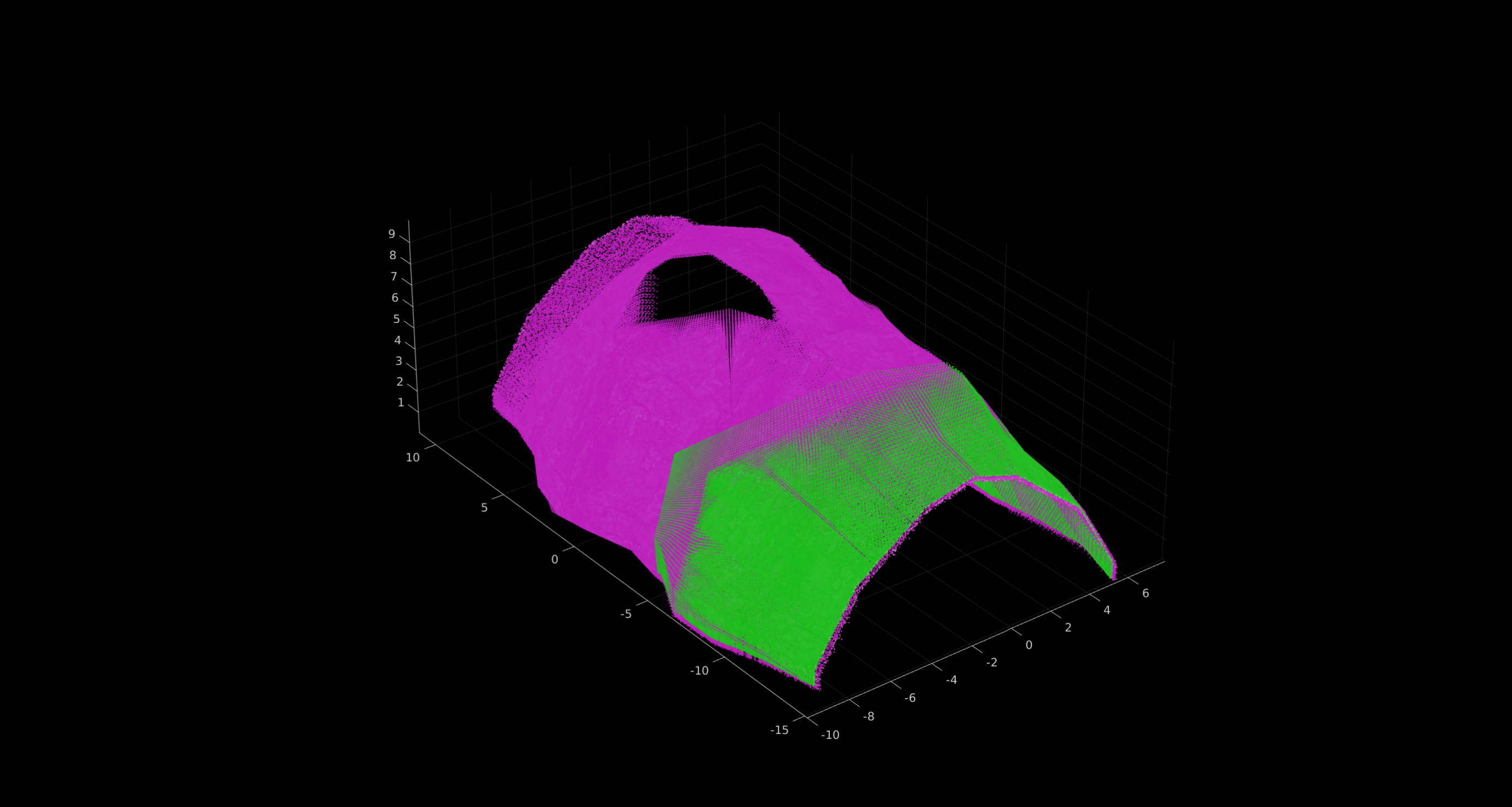} \\
        \textbf{Pointcloud 4 (Before Registration)} & \textbf{Pointcloud 4 (After Registration)} \\
    \end{tabular}
    \caption{Pointclouds before and after registration using the full framework. The source pointcloud is in magenta and the target pointcloud is in green. }
    \label{reg_gazebo}
\end{figure*}

To validate the framework both simulation and real-world tests were conducted. The ROS/Gazebo\cite{takaya2016simulation} computational simulator with the model mine environment shown in Figure \ref{gazebo_mine}, a drone fitted with a depth camera and RTABMap\cite{labbe2019rtab}, a widely used open-source SLAM (Simultaneous Localization and Mapping) framework were used to generate maps of the model environment. A ground-truth point cloud map of the environment and four additional point clouds but from different starting poses were generated. In order to provide insights on the proposed approach, the framework was evaluated by registering each of the four pointclouds to the ground-truth map using different algorithmic configurations: FPFH alone, FPFH combined with NDT, FPFH combined with ICP, NDT combined with ICP, and the full pipeline integrating FPFH, NDT, and ICP. The registration performance was assessed using the inlier percentage of the registered points and the root mean square error (RMSE) of the inlier points as shown in Table \ref{tab:registration_results}. A point in the target pointcloud is considered an inlier if the distance to its nearest neighbor in the source pointcloud is less than 0.5m. The 0.5m threshold was chosen as a conservative measure to ensure the RMSE reflects meaningful alignment accuracy. A lower threshold could exclude valid correspondences, leading to artificially low RMSE values even in poor registrations. Figure \ref{reg_gazebo} visually presents the registration results obtained using the full pipeline for each point cloud.

\begin{table*}[h]
    \centering
    \caption{Registration performance for different framework configurations for Gazebo simulation environment}
    \label{tab:registration_results}
    \begin{tabular}{|l|cc|cc|cc|cc|}
        \hline
        \multirow{2}{*}{\textbf{Framework Configuration}} & \multicolumn{2}{c|}{\textbf{Pointcloud 1}} & \multicolumn{2}{c|}{\textbf{Pointcloud 2}} & \multicolumn{2}{c|}{\textbf{Pointcloud 3}} & \multicolumn{2}{c|}{\textbf{Pointcloud 4}} \\
        \cline{2-9}
        & \textbf{RMSE (m)} & \textbf{Inlier \%} & \textbf{RMSE (m)} & \textbf{Inlier \%} & \textbf{RMSE (m)} & \textbf{Inlier \%} & \textbf{RMSE (m)} & \textbf{Inlier \%} \\
        \hline
        FPFH Only                        & 0.1253 & 86.59 & 0.1068 & 90.55 & 0.2722 & 33.97 & 0.2472 & 78.26 \\
        \hline
        FPFH + NDT                        & 0.0988 & 88.80 & 0.0909 & 89.32 & 0.1504 & 87.61 & 0.1365 & 89.35 \\
        \hline
        FPFH + ICP                        & 0.1231 & 87.94 & 0.0993 & 89.94 & 0.2329 & 57.28 & 0.2394 & 88.12 \\
        \hline
        NDT + ICP                        & 0.2432 & 70.00 & 0.1965 & 77.78 & 0.2817 & 43.03 & 0.1938 & 68.51 \\
        \hline
        FPFH + NDT + ICP                   & 0.1114 & 89.11 & 0.1324 & 87.21 & 0.1226 & 87.80 & 0.1820 & 89.33 \\
        \hline
    \end{tabular} \\
    \vspace{2mm}
    \footnotesize{RMSE values are in meters, and inlier percentages are shown as \%.}
\end{table*}

The results showcase the effectiveness of the full registration framework, which balances performance across different scenarios. Although it does not always achieve the lowest RMSE or the highest inlier percentage, it reliably delivers consistent registration across the range of cases. For example, in Pointcloud 3, the full pipeline achieves an RMSE of 0.1226 m and an inlier percentage of 87.80\%, successfully addressing that difficult initial misalignment. The use of FPFH alone produces reasonable initial alignments but struggles with more challenging pointclouds. For example, in Pointcloud 3, FPFH achieves an RMSE of 0.2722 m with only 33.97\% inliers. Combining FPFH with NDT significantly improves the results, as evidenced by the reduced RMSE and increased inlier percentages across all point clouds. FPFH combined with ICP also shows notable improvements over FPFH alone but does not perform as well as FPFH + NDT in some cases. In challenging cases like Pointcloud 3, FPFH + ICP achieves an RMSE of 0.2329 m, which is higher than the RMSE achieved by FPFH + NDT. The worst results were with NDT + ICP indicating the importance of the initialization provided by FPFH. It should be noted that the target pointcloud can contain regions that are not in the source pointcloud and in such a case, the inlier percentage cannot be up to 100\%. However, this metric remains valid for relative performance evaluations across methods, as this limitation applies equally to all tests.

\begin{figure*}[h!]
    \centering
    \begin{tabular}{cc}
        \includegraphics[width=0.49\textwidth, height=0.35\textwidth]{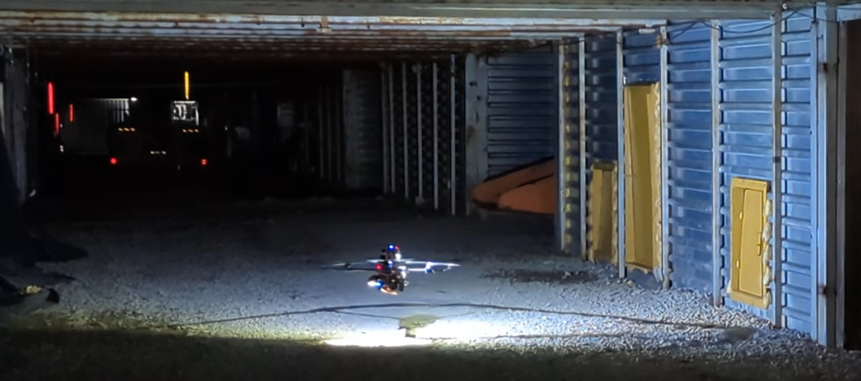} & 
        \includegraphics[width=0.49\textwidth, height=0.35\textwidth]{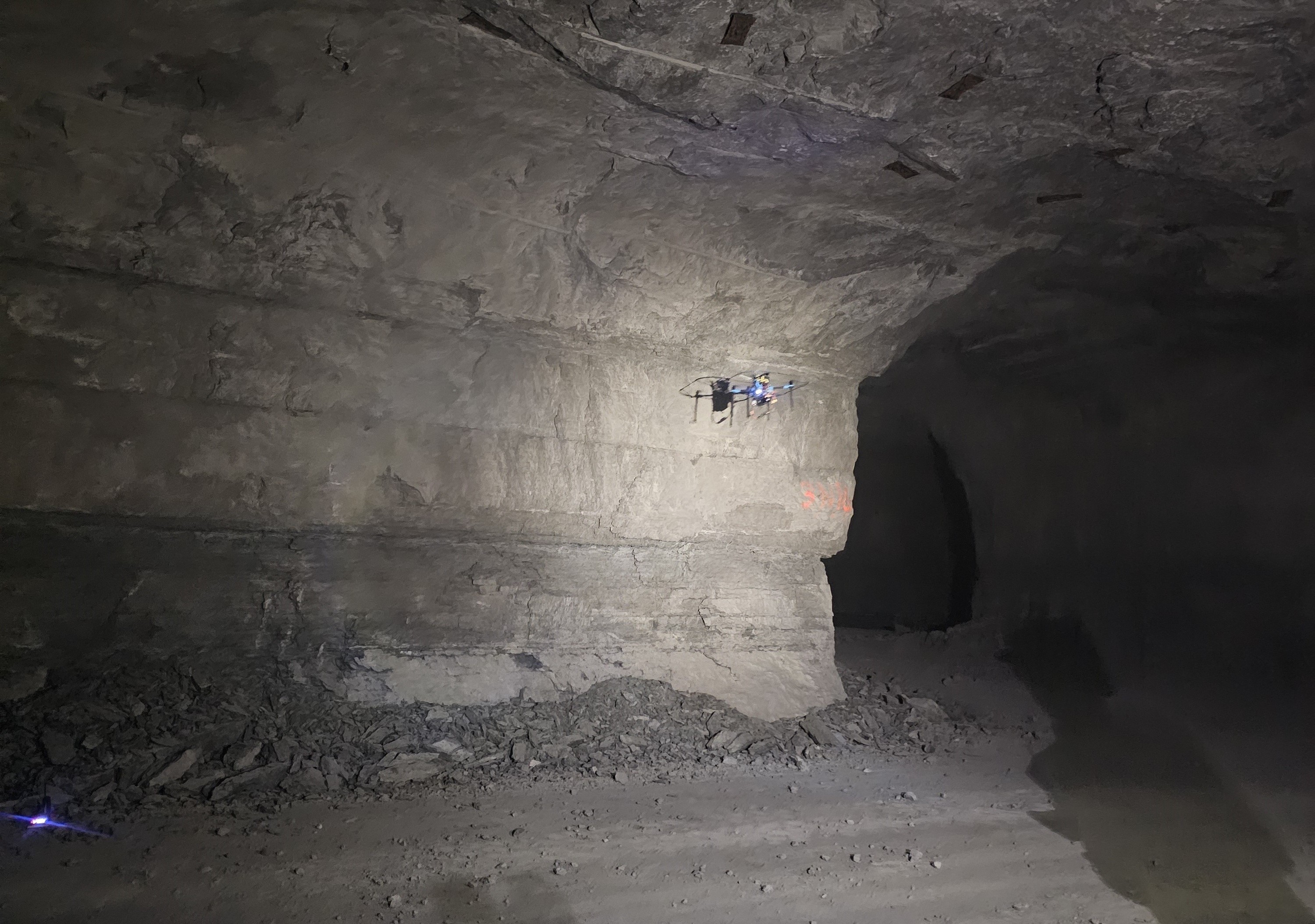} \\
        \textbf{Simulation Mine} & \textbf{Limestone Mine} \\
    \end{tabular}
    \caption{Both test environments: the physically simulated coalmine (left) and the actual limestone mine (right).}
    \label{real-world}
\end{figure*}

The framework was also tested in two different real-world settings: one in a controlled physically simulated coalmine, designed to replicate mine-like conditions, and the other in an actual limestone mine. Both environments are shown in Figure \ref{real-world}, with the physically simulated coalmine providing a controlled setting for systematic testing, while the limestone mine presented real-world challenges for evaluating the framework under dynamic and unpredictable conditions.

In the physically simulated coalmine, camera image data was collected from one of the walls while flying a drone equipped with an Intel RealSense T265 stereo camera and an Intel RealSense LiDAR Camera L515 \cite{martinez2023oxpecker}. Using the collected data and RTABMap\cite{labbe2019rtab}, used to generate two distinct maps, shown in Figure \ref{registration_results_real}. In the limestone mine, camera image data was also collected from the lower portion of a mine wall while autonomously flying the same drone, this time equipped with an Intel RealSense D457 depth camera. The data collected was used to generate two separate maps, as shown in Figure \ref{registration_results_real}.

\begin{figure*}[h!]
    \centering
    \begin{tabular}{cc}
        \includegraphics[width=0.49\textwidth, clip, trim=300 100 300 100]{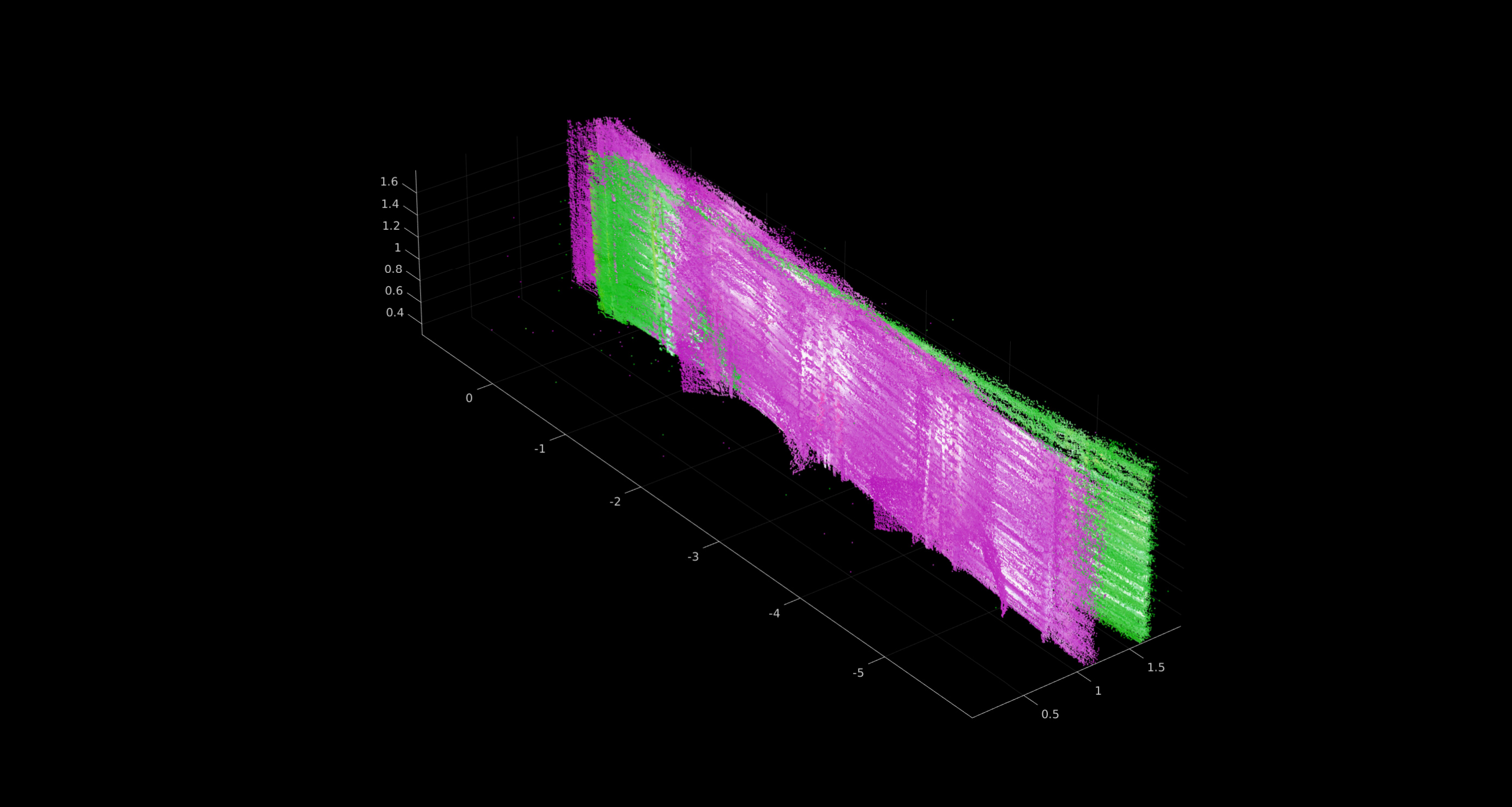} & 
        \includegraphics[width=0.49\textwidth, clip, trim=300 100 300 100]{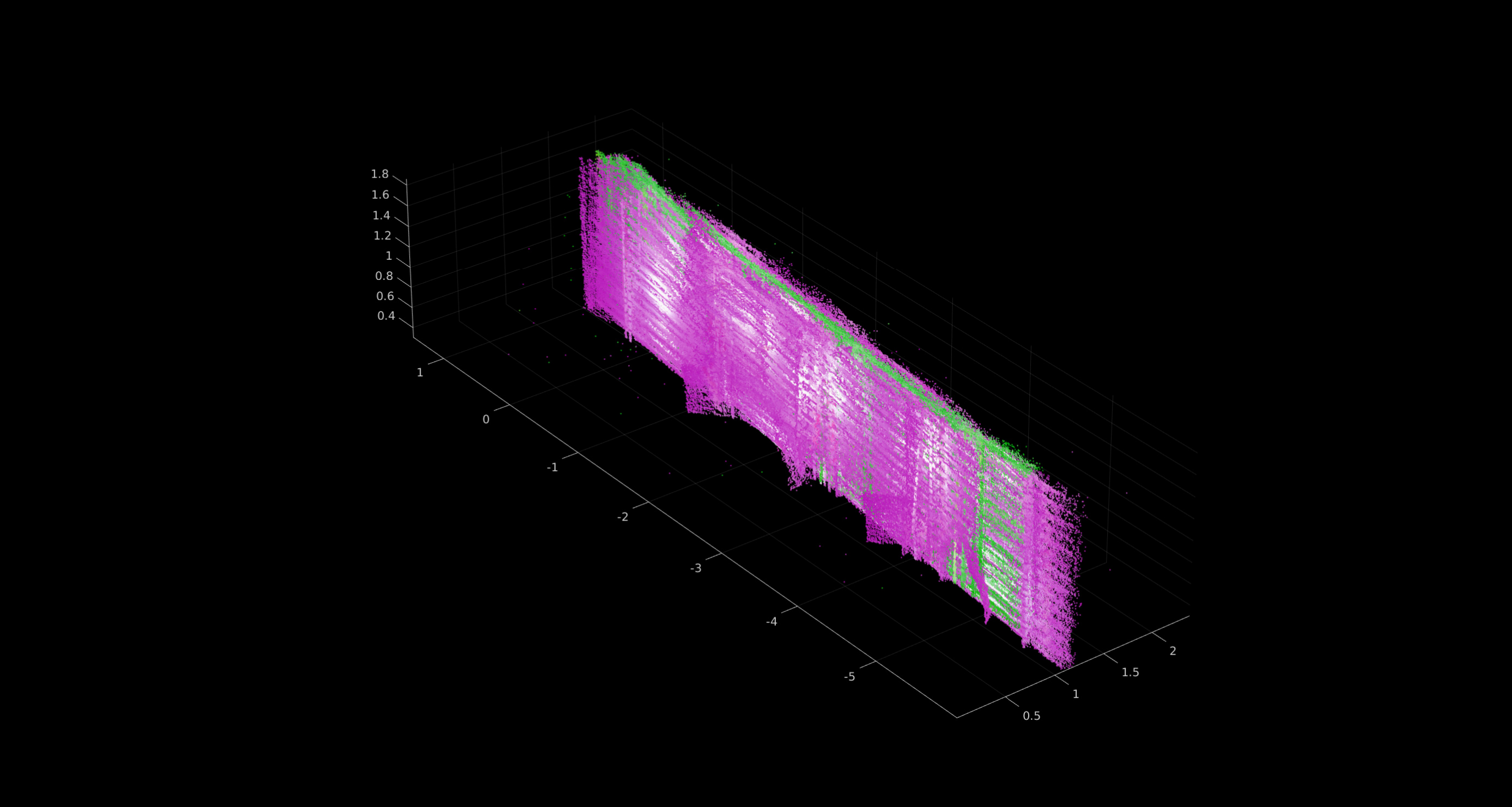} \\
        \textbf{Before Registration - Simulation Mine} & \textbf{After Registration - Simulation Mine} \\
        \includegraphics[width=0.49\textwidth, clip, trim=300 100 300 100]{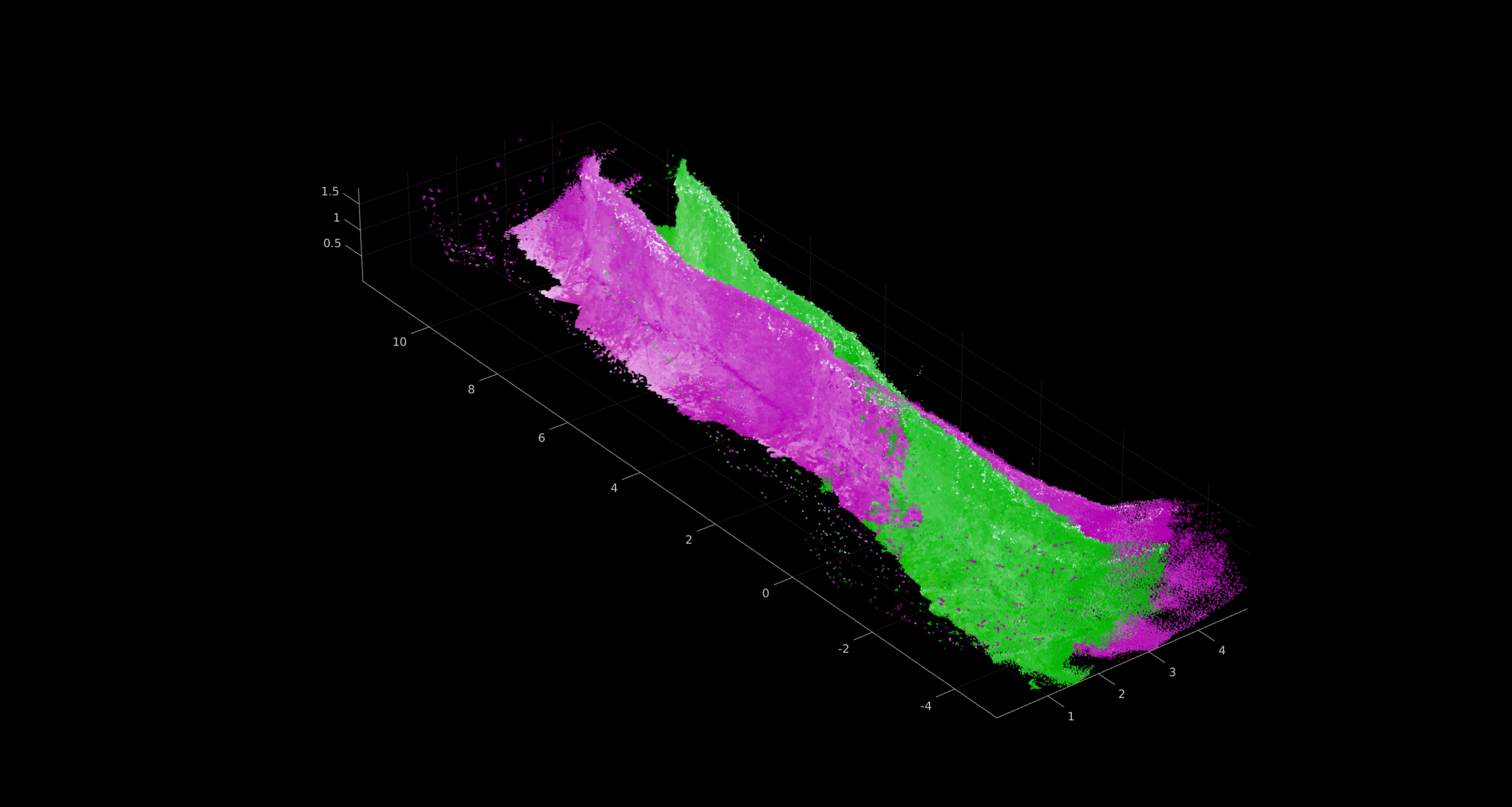} & 
        \includegraphics[width=0.49\textwidth, clip, trim=300 100 300 100]{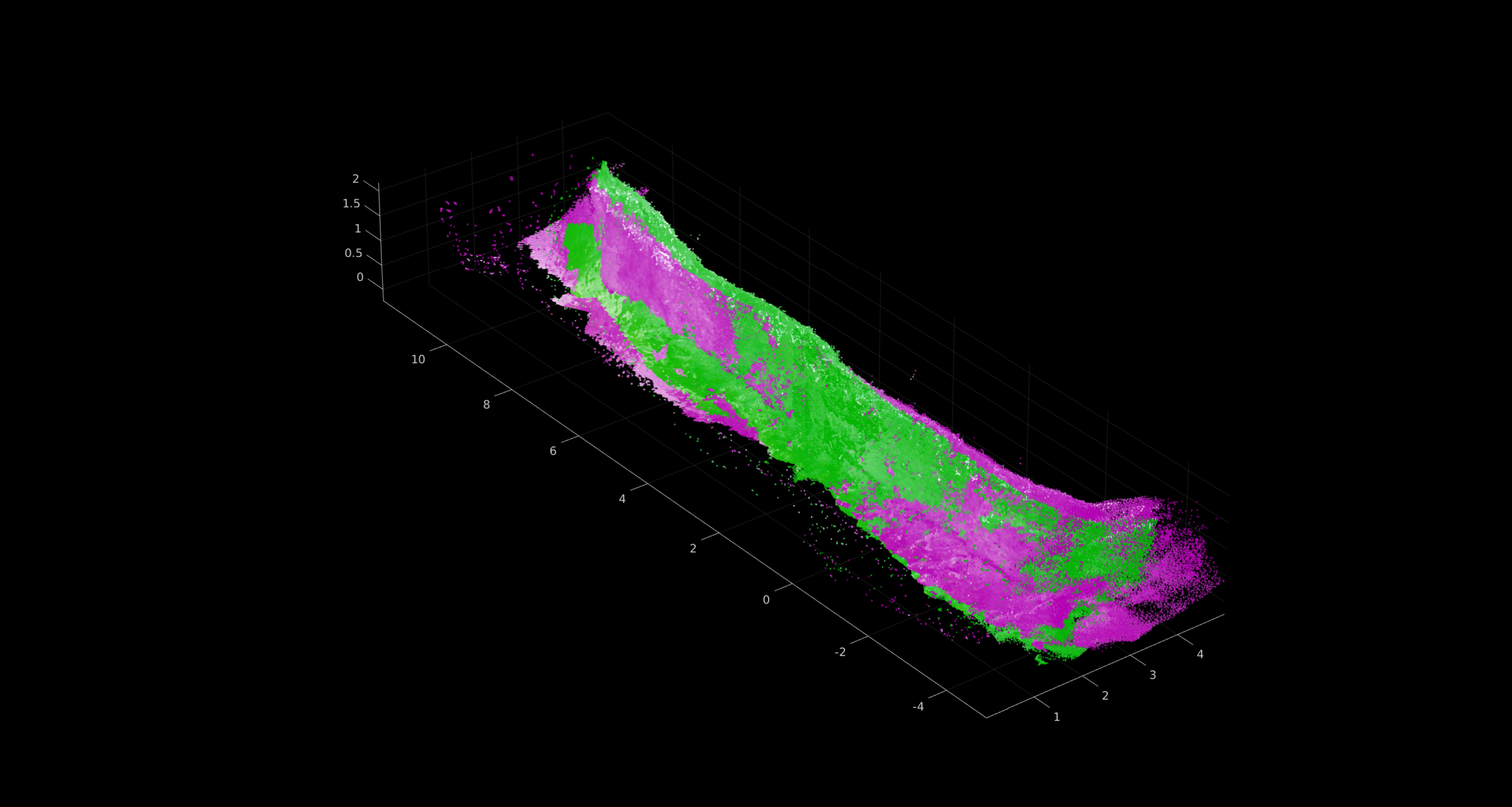} \\
        \textbf{Before Registration - Limestone Mine} & \textbf{After Registration - Limestone Mine} \\
    \end{tabular}
    \caption{Point cloud registration results: (Top) Simulation mine before and after registration. (Bottom) Limestone mine before and after registration. The source pointcloud is in magenta and the target pointcloud is in green.}
    \label{reg_real}
\end{figure*}

Similar to the Gazebo computer-based simulation, the framework was evaluated in both environments by registering each point cloud using various algorithmic configurations: FPFH alone, FPFH combined with NDT, FPFH combined with ICP, NDT combined with ICP, and the full pipeline integrating FPFH, NDT, and ICP. The performance of the registration was assessed using two metrics: the root mean square error (RMSE) of the registered points and the inlier percentage, as presented in Table \ref{registration_results_real}. The registration results using the full pipeline for each point cloud are visually presented in Figure \ref{reg_real}

\begin{table*}[h]
    \centering
     \caption{Registration performance in a physically simulated coalmine and real limestone mine using different algorithmic configurations. RMSE and inlier percentage metrics are presented.}
     \label{registration_results_real}
     \begin{tabular}{|l|cc|cc|}
        \hline
        \multirow{2}{*}{\textbf{Framework Configuration}} & \multicolumn{2}{c|}{\textbf{Simulation Mine}} & \multicolumn{2}{c|}{\textbf{Limestone Mine}} \\
        \cline{2-5}
        & \textbf{RMSE (m)} & \textbf{Inlier \%} & \textbf{RMSE (m)} & \textbf{Inlier \%}\\
        \hline
        FPFH Only                        & 0.1325 & 99.99 & 0.0703 & 99.96 \\
        \hline
        FPFH + NDT                        & 0.0173 & 99.97 & 0.0751 & 99.98 \\
        \hline
        FPFH + ICP                        & 0.0196 & 100.00 & 0.0651 & 99.98 \\
        \hline
        NDT + ICP                        & 0.0403 & 99.96 & 0.1398 & 98.90 \\
        \hline
        FPFH + NDT + ICP                   & 0.0173 & 99.97 & 0.0941 & 99.91 \\
        \hline
    \end{tabular} \\
    \vspace{2mm}
    \footnotesize{RMSE values are in meters, and inlier percentages are shown as \%.}
\end{table*}

The framework performed well in the simulation mine as well as in the limestone mine, showing its robustness in diverse conditions. In the simulation mine, the best performance was achieved with FPFH + ICP, which had an RMSE of 0.0196m and 100\% inliers. FPFH + NDT also performed well, with an RMSE of 0.0173m and 99.97\% inliers. The full pipeline (FPFH + NDT + ICP) did not show a significant improvement over these configurations. In the limestone mine, FPFH alone had an RMSE of 0.0703m and 99.96\% inliers. More sophisticated configurations, such as FPFH + ICP and FPFH + NDT produced better results with their respective RMSEs being 0.0651m and 0.0751m. The complete pipeline obtained a marginally higher RMSE of 0.0941m and 99.91\% inliers. In general, simpler configurations like FPFH + ICP or FPFH + NDT performed sufficiently well while the full pipeline maintained consistent performance and was responsive to real-world condition difficulties.

\section{Conclusion}
\label{conclusion}
The goal of this work was to develop and evaluate an effective relocalization framework that is more robust under challenging underground conditions by addressing issues of noise, occlusions, and irregular surfaces while maintaining computational speed for real-time applications in autonomous robots. The proposed framework showed potential as it was successfully tested on both computer simulated, physically simulated mines, and real mine datasets. The integration of Intrinsic Shape Signatures (ISS)\cite{zhong2009intrinsic} for keypoints detection, Fast Point Feature Histogram (FPFH) \cite{rusu2009fast} for matching descriptors, and two-stage transformation refinement using Normal Distributions Transform (NDT)\cite{magnusson2007scan} and Iterative Closest Point (ICP)\cite{pomerleau2013comparing} registration improved the robustness and accuracy of relocalization even in noisy situations and challenging initial conditions. The performance of the full proposed framework was consistently balanced across the different tested scenarios. Even though it did not always give the lowest RMSE or highest inlier percentage, it provided reliable registration even in challenging conditions. For instance, it decreased the RMSE to roughly 0.12 m with nearly 88\% inliers in Pointcloud 3 of the computer simulation, as opposed to 0.27 m RMSE and 34\% inliers for FPFH alone. The results highlight the impact of initialization—FPFH alone provided reasonable alignments but struggled with difficult cases. Adding NDT lowered the RMSE, outperforming FPFH + ICP in some cases. The poorest results were with NDT + ICP, emphasizing the importance of feature-based initialization. Future work will focus on expanding dataset collection from various subterranean environments and conditions (e.g., dust, moisture) to further refine our approach and optimize algorithms for feature extraction, point cloud registration, and transformation estimation to improve performance and reliability. Additionally, we aim to explore the integration of the framework with autonomous robotic systems to assess its performance in real-time navigation tasks and decision-making processes in underground settings. The framework’s demonstrated effectiveness provides a strong foundation for its potential deployment in critical applications like exploration, search and rescue, and safety inspections in underground environments.

\section*{Acknowledgements}
This research was sponsored in part by the Alpha Foundation awarded AFCTG22R2-159 entitled ``Design Guidelines for Assessment of Pillar Stability in Underground Room \& Pillar Mines from Autonomous Robotic Inspections” and by the Statler College of Engineering and Mineral Resources Christopher PhD Fellowship. The authors would like to thanks Dr. Guilherme Periera and Mr. Luis Escobar of the WVU Field and Aerial Robotics Lab (FARO) lab for their support of the drone pointcloud data collection. 

\bibliographystyle{IEEEtran}
\bibliography{references}

\end{document}